%% file: main.tex
\documentclass[conference]{IEEEtran}
\usepackage{todonotes}

\usepackage{tabularray}
\usepackage{graphicx}
\usepackage{float}
\usepackage{amsmath,amssymb,amsfonts}
\usepackage{algorithmic}
\usepackage{textcomp}
\usepackage{xcolor}
\usepackage{adjustbox}
\usepackage{float}
\usepackage[utf8]{inputenc}
\usepackage{caption}
\usepackage{comment}
\usepackage{wrapfig}
\usepackage[compact]{titlesec}
\usepackage[colorlinks=true,urlcolor=black]{hyperref}   

\AtBeginDocument{%
  \providecommand\BibTeX{{%
    \normalfont B\kern-0.5em{\scshape i\kern-0.25em b}\kern-0.8em\TeX}}}

\IEEEoverridecommandlockouts
\begin{document}

\title{Advanced Brain Tumor Segmentation Using
EMCAD: Efficient Multi-scale Convolutional Attention Decoding}

\makeatletter 
\newcommand{\linebreakand}{%
  \end{@IEEEauthorhalign}
  \hfill\mbox{}\par
  \mbox{}\hfill\begin{@IEEEauthorhalign}
}
\makeatother 

\author{

\IEEEauthorblockN{GodsGift Uzor}
\IEEEauthorblockA{\textit{Department of Computer Science} \\
\textit{Texas Tech University}\\
Lubbock, TX \\
godsgift.uzor@ttu.edu}

\and
\IEEEauthorblockN{Tania-Amanda Nkoyo Fredrick Eneye}
\IEEEauthorblockA{\textit{Department of Computer Science} \\
\textit{Texas Tech University}\\
Lubbock, TX \\
tafredri@ttu.edu}

\linebreakand
\and
\IEEEauthorblockN{Ijezue Chukwuebuka}
\IEEEauthorblockA{\textit{Department of Computer Science} \\
\textit{Texas Tech University}\\
Lubbock, TX  \\
cijezue@ttu.edu}
}






\maketitle


\input{sections/abstract}



\input{sections/introduction}
\input{sections/related-work}
\input{sections/emcad_methodology}

\input{sections/implementation}

\input{sections/future-work}

\input{sections/conclusion}

\bibliographystyle{IEEEtran}
\bibliography{./bib/main}

\end{document}

%% file: sections/abstract.tex
\begin{abstract}


Brain tumor segmentation is a critical pre-processing step in the medical image analysis pipeline that involves precise delineation of tumor regions from healthy brain tissue in medical imaging data, particularly MRI scans. An efficient and effective decoding mechanism is crucial in brain tumor segmentation especially in scenarios with limited computational resources. However these decoding mechanisms usually come with high computational costs. To address this concern EMCAD a new efficient multi-scale convolutional attention decoder designed was utilized to optimize both performance and computational efficiency for brain tumor segmentation on the BraTs2020 dataset consisting of MRI scans from 369 brain tumor patients. 
The preliminary result obtained by the model achieved a best Dice score of 0.31 and maintained a stable mean Dice score of 0.285 ± 0.015 throughout the training process which is moderate. The initial model maintained consistent performance across the validation set without showing signs of over-fitting.
 
\end{abstract}

%% file: sections/introduction.tex
\section{Introduction}
\subsection{Medical Image Segmentation}
Medical image segmentation is a crucial process in medical image analysis that involves partitioning medical images into multiple meaningful segments or regions, each corresponding to different anatomical structures, tissues, or pathologies~\cite{shen2017deep}. This computational technique has evolved significantly with the advent of deep learning approaches, enabling automatic delineation of regions of interest from various imaging modalities such as MRI, CT, and ultrasound~\cite{shin2016deep}. The segmentation process helps in extracting quantitative information from medical images, which is essential for diagnosis, treatment planning, and follow-up assessment~\cite{warfield2004simultaneous}. Traditional segmentation methods relied heavily on handcrafted features and manual intervention, but modern deep learning-based approaches have revolutionized this field by offering more robust, accurate, and automated solutions~\cite{litjens2017survey}. These advanced techniques can automatically learn hierarchical feature representations from raw image data, leading to more precise and reliable segmentation results across different medical imaging applications~\cite{wang2018interactive}.

Medical image segmentation enables precise delineation of anatomical structures, pathological regions, and abnormalities, which is essential for accurate diagnosis, volumetric analysis, and disease progression monitoring~\cite{shen2017deep}. In clinical practice, segmentation algorithms assist healthcare professionals in detecting and analyzing various medical conditions, including tumors, lesions, and organ abnormalities, leading to improved diagnostic accuracy and treatment outcomes~\cite{litjens2017survey}. Furthermore, automated segmentation techniques significantly reduce the time-consuming nature of manual annotation while maintaining high precision, thereby enhancing workflow efficiency in clinical settings~\cite{tajbakhsh2020embracing}. The increasing adoption of artificial intelligence in healthcare has made medical image segmentation even more critical, as it serves as a preprocessing step for various downstream tasks such as disease classification, prognosis prediction, and treatment response assessment~\cite{zhou2019review}.
\subsection{Brain Tumor Segmentation}
Brain tumor segmentation is a critical preprocessing step in the medical image analysis pipeline that involves precise delineation of tumor regions from healthy brain tissue in medical imaging data, particularly MRI scans~\cite{litjens2017survey}. This process encompasses the identification and isolation of various tumor components, including active tumor regions, necrotic cores, and surrounding edema~\cite{bauer2013survey}. The complexity of brain tumor segmentation stems from the heterogeneous nature of tumor appearances, varying locations, diverse shapes, and ambiguous boundaries that often resemble normal brain tissue~\cite{kuijf2019standardized}. As manual segmentation is time-consuming and subject to inter-observer variability, automated and semi-automated segmentation methods have become increasingly important in clinical practice.

Brain tumor segmentation plays a fundamental role in clinical practice, serving as a crucial tool for diagnosis, treatment planning, and monitoring disease progression~\cite{menze2014multimodal}. Accurate segmentation enables volumetric analysis of tumors, which is essential for evaluating treatment response and adjusting therapeutic strategies. Gab Allah et al.~\cite{allah2023edge}demonstrated that precise tumor boundary delineation significantly impacts treatment planning and surgical approach selection. Moreover, accurate segmentation facilitates better understanding of tumor growth patterns and infiltration into surrounding tissues, which is crucial for predicting patient outcomes and developing personalized treatment strategies. The segmentation results also serve as valuable input for radiotherapy planning, helping to minimize damage to healthy tissue while maximizing therapeutic effect on tumor regions.

The advent of neural networks has revolutionized brain tumor segmentation, offering superior performance compared to traditional image processing methods. Deep learning approaches, particularly convolutional neural networks (CNNs), have demonstrated remarkable ability to learn hierarchical features directly from medical imaging data, leading to more robust and accurate segmentation results~\cite{pereira2016brain}. Recent studies have shown that attention-based architectures can achieve Dice scores exceeding 90\% for various tumor types~\cite{alirr2023automatic}, while efficient architectures like QuickTumorNet have demonstrated the potential for real-time segmentation with clinically acceptable accuracy~\cite{maas2021quicktumornet}. These advancements have not only improved segmentation accuracy but also enhanced the efficiency of clinical workflows, enabling faster and more reliable tumor analysis in medical practice~\cite{almufareh2024automated}.

\subsection{Introducing EMCAD}
\begin{figure}[!h]
    \centering
    \includegraphics[width=\linewidth]{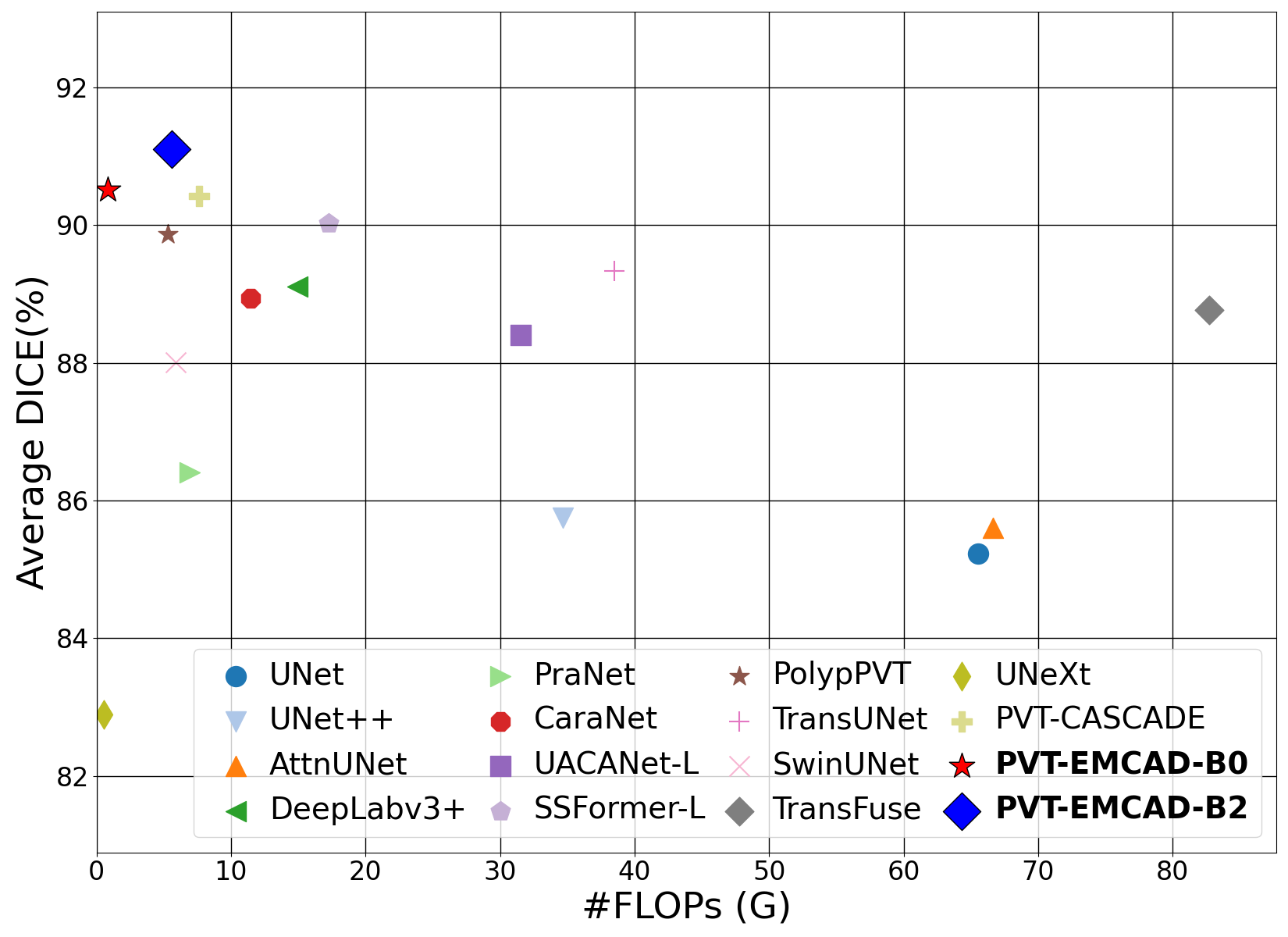}
    \caption{Comparing EMCAD with other state of the art models \cite{rahman2024emcad}}
    \label{fig:emcad_sota_comparison}
\end{figure}

The EMCAD architecture is an efficient, lightweight model optimized for 2D medical image segmentation, balancing high accuracy with low computational cost. It employs a multi-scale depth-wise convolution block (MSCB) with parallel kernel sizes ($3 \times 3$, $5 \times 5$, and $7 \times 7$) to capture complex patterns, enhancing feature representation with minimal resources. The efficient multi-scale convolutional attention module (EMCAM) selectively refines features from the encoder, isolating critical areas by suppressing irrelevant regions. EMCAD also integrates a large-kernel grouped attention gate (LGAG), which fuses refined features using $3 \times 3$ grouped convolutions, enhancing the contextual understanding of salient regions. With 0.506M parameters and 0.11 GFLOPs for its tiny encoder configuration, EMCAD achieves superior segmentation performance, outperforming state-of-the-art models with reduced computational demand see figure \ref{fig:emcad_sota_comparison}

%% file: sections/related-work.tex
\section{Related Work}
\subsection{\textbf{Vision Encoders}}
Convolutional Neural Networks (CNNs) have served as foundational architectures for medical image analysis due to their ability to effectively handle spatial relationships. AlexNet~\cite{NIPS2012_c399862d} and VGG~\cite{simonyan2015deepconvolutionalnetworkslargescale} pioneered deep convolutional architectures for feature extraction. GoogleNet~\cite{Szegedy_2015_CVPR} introduced the inception module to enable more efficient multi-scale feature computation. A significant advancement came with ResNet~\cite{He_2016_CVPR}, which addressed the vanishing gradient problem through residual connections, allowing for substantially deeper networks. MobileNets~\cite{howard2017mobilenets} brought CNNs to resource-constrained environments through lightweight depth-wise separable convolutions.

More recently, Vision Transformers (ViTs) have emerged as powerful alternatives to CNNs, pioneered by Dosovitskiy et al.'s work~\cite{dosovitskiy2020image}. ViTs excels at capturing long-range dependencies through self-attention mechanisms. Subsequent developments like Swin Transformer~\cite{liu2021swin} incorporated sliding window attention, while PVT~\cite{wang2021pyramid} introduced spatial reduction attention for improved efficiency. These architectures have shown particular promise in medical imaging applications due to their ability to model complex global relationships in anatomical structures.

\subsection{\textbf{Medical Image Segmentation}}
Medical image segmentation has witnessed significant architectural evolution, driven by the need for both accuracy and computational efficiency in clinical settings. Early approaches relied heavily on traditional image processing techniques, but the field was revolutionized with the advent of deep learning architectures. U-shaped architectures emerged as the cornerstone of medical image segmentation, with the original U-Net~\cite{ronneberger2015u} establishing the encoder-decoder paradigm with skip connections. This fundamental architecture enabled precise localization while maintaining contextual information, setting a new standard for medical image segmentation tasks.

The field has since progressed along three main directions: architectural efficiency, attention mechanisms, and multi-scale processing. UNet++\cite{zhou2018unet++} enhanced the basic U-Net design through nested, dense skip connections, while DeepLab variants~\cite{chen2017deeplab} introduced atrous convolutions and spatial pyramid pooling for better multi-scale handling. The advent of nnU-Net~\cite{isensee2021nnu} demonstrated the potential of automated architecture adaptation, configuring network parameters based on dataset characteristics. Parallel to these developments, attention mechanisms have become increasingly prominent, with architectures like PraNet~\cite{fan2020pranet} utilizing reverse attention for feature refinement, and CASCADE~\cite{rahman2023medical} implementing cascaded attention in decoder frameworks, significantly improving the capture of fine anatomical details and boundary delineation.

\subsubsection{Efficient Decoder Architectures}
Efficient decoder architectures have emerged as a critical research direction in medical image segmentation, addressing the computational demands of traditional architectures. Mehta et al.~\cite{mehta2018espnet} introduced ESPNet, which utilized efficient spatial pyramid convolutions in the decoder, achieving comparable accuracy to larger networks while requiring significantly fewer parameters (approximately 0.4M). Building on this efficiency-focused approach, Paszke et al.~\cite{paszke2016enet} proposed ENet, demonstrating that carefully designed decoder architectures could maintain high accuracy while reducing computational complexity by up to 18x compared to traditional methods. Yu et al.~\cite{yu2018bisenet} further advanced this field with BiSeNet, introducing a bilateral segmentation network that balanced the need for spatial details and contextual information through a dual-path decoder structure, achieving real-time performance while maintaining competitive accuracy on medical imaging tasks.

\subsubsection{Advanced Attention Mechanisms}
The integration of attention mechanisms in medical image segmentation has evolved significantly, moving beyond simple spatial attention to more sophisticated approaches that enhance feature representation while maintaining computational efficiency. Oktay et al.~\cite{oktay2018attention} introduced the Attention U-Net, pioneering the use of attention gates in medical image segmentation. Their approach demonstrated that attention mechanisms could automatically learn to focus on target structures of varying shapes and sizes, particularly beneficial for pancreatic segmentation where precise localization is crucial.
Subsequent developments saw the emergence of dual attention networks, exemplified by Fu et al.\cite{fu2019dual}, who proposed DANet to adaptively integrate local features with their global dependencies. By introducing parallel position and channel attention modules, DANet captured both spatial and channel dependencies, significantly improving segmentation performance in medical imaging tasks. This dual attention approach proved particularly effective in capturing long-range dependencies and highlighting diagnostically relevant regions.
The transformer revolution in computer vision sparked a new wave of attention-based architectures. Chen et al.\cite{chen2021transunet} proposed TransUNet, effectively combining transformers with U-Net architecture for medical image segmentation. Their hybrid approach leveraged the global context modeling capability of transformers while maintaining the detailed local feature processing of CNNs, achieving state-of-the-art performance on various medical segmentation benchmarks. The success of TransUNet demonstrated that transformer-based attention could effectively capture long-range dependencies crucial for accurate medical image segmentation.
Recent attention mechanism developments have focused on efficiency and scalability. Wang et al.\cite{wang2021pyramid} introduced the Pyramid Vision Transformer (PVT), which incorporated a progressive shrinking pyramid and spatial-reduction attention to reduce computational complexity while maintaining performance. This advancement was particularly significant for medical image segmentation, where high-resolution inputs are common and computational efficiency is crucial. Similarly, Liu et al.\cite{liu2021swin} proposed Swin Transformer, introducing shifted windows for more efficient attention computation, proving especially effective in handling varying tumor sizes and shapes in medical imaging.

\subsubsection{Multi-scale Approaches}
Multi-scale processing has become increasingly crucial in medical image segmentation, particularly for handling varying anatomical structures and pathological features at different scales. Chen et al.\cite{chen2017deeplab} pioneered the effective use of multi-scale context in deep networks through DeepLab, demonstrating that atrous spatial pyramid pooling could systematically capture information at multiple scales. Their approach effectively balanced field-of-view and computational efficiency, providing a foundation for modern multi-scale architectures.
The Feature Pyramid Network (FPN) introduced by Lin et al.\cite{lin2017feature} revolutionized multi-scale feature processing by creating a feature hierarchy with lateral connections, effectively combining low-resolution, semantically strong features with high-resolution, semantically weak features. This architecture proved particularly valuable in medical image segmentation, where preserving both fine details and global context is essential for accurate boundary delineation.
Wang et al.\cite{wang2020deep} advanced this concept further with High-Resolution Networks (HRNet), maintaining high-resolution representations throughout the network while creating reliable high-resolution-to-low-resolution connections. HRNet's ability to maintain high-resolution features throughout the network proved especially beneficial for medical image segmentation tasks where precise spatial information is crucial. The parallel multi-resolution subnetworks and repeated multi-scale fusions throughout the network enabled better feature representation at multiple scales.
Recent developments in multi-scale approaches have focused on efficient scale handling in medical contexts. Zhou et al.\cite{zhou2019unet++} demonstrated the effectiveness of multi-scale feature fusion in UNet++, introducing dense skip connections to bridge the semantic gap between encoder and decoder features. Similarly, Isensee et al.\cite{isensee2021nnu} showed through nnU-Net that carefully designed multi-scale architectures could achieve state-of-the-art performance across diverse medical segmentation tasks without task-specific adaptations.

\subsection{\textbf{Brain Tumor Segmentation}}
Brain tumor segmentation represents a particularly challenging domain within medical image segmentation due to the heterogeneous nature of tumors, varying sizes, irregular shapes, and complex infiltration patterns. The field has evolved significantly with the advent of deep learning, progressing along three main trajectories: architectural innovations for improved accuracy, computational efficiency for clinical deployment, and multi-scale processing for handling tumor variability.

Recent architectural innovations have demonstrated significant improvements in segmentation accuracy. Edge U-Net\cite{allah2023edge} marked a notable advancement by incorporating boundary information through an innovative Edge Guidance Block (EGB), achieving impressive Dice scores of 88.8\%, 91.8\%, and 87.3\% for meningioma, glioma, and pituitary tumors respectively. This boundary-aware approach was further enhanced by Alirr et al.\cite{alirr2023automatic}, who integrated a convolutional block attention module (CBAM) with edge-enhancing diffusion preprocessing, showing superior performance on the BRATS 2020 dataset.

The push toward clinical applicability has driven developments in computational efficiency. QuickTumorNet\cite{maas2021quicktumornet} exemplified this trend, achieving practical inference speeds of 100-143 FPS while maintaining clinically viable accuracy (Dice scores of 64\% for glioma, 80.1\% for meningioma, and 81.1\% for pituitary tumors). Similarly, YOLO-based approaches by Almufareh et al.\cite{almufareh2024automated} demonstrated the potential of object detection frameworks, achieving high precision (93.6\%) and recall (90.6\%) while maintaining computational efficiency.

Hybrid approaches have also emerged, combining traditional and deep learning methods. Notably, Rajendran et al.\cite{rajendran2023automated} integrated Gray Level Co-occurrence Matrix (GLCM) features with deep learning, achieving comprehensive performance metrics above 98\% for whole tumor, enhanced tumor, and tumor core segmentation. These developments highlight the field's evolution toward more accurate, efficient, and clinically applicable methods.

\subsubsection{Computational Efficiency}
Ghaffari et al.\cite{ghaffari2019automated} advanced computational efficiency through a lightweight CNN architecture specifically designed for real-time brain tumor segmentation. Their model achieved significant reduction in computational complexity while maintaining a competitive Dice score of 0.85 on the BraTS dataset, proving particularly effective for resource-constrained clinical environments.
Pereira et al.\cite{pereira2016brain} presented an efficient architecture using small kernels in CNN design, which significantly reduced the computational complexity while achieving robust segmentation results across different tumor types. Their approach demonstrated that carefully designed architectures could maintain high accuracy while reducing computational demands.
Dong et al/\cite{dong2017automatic} introduced U-Net-based lightweight architecture that achieved efficient brain tumor segmentation through channel pruning and knowledge distillation. Their method substantially reduced model parameters while maintaining competitive performance on the BraTS dataset, making it suitable for resource-constrained environments.
Huang et al.\cite{huang2014brain} proposed an efficient approach based on local independent projection, demonstrating that well-designed feature extraction methods could achieve accurate tumor segmentation while maintaining computational efficiency. Their method showed particular effectiveness in handling the computational challenges of processing multi-modal MRI data.

\subsubsection{Multi-scale Processing for Tumors subsection}
Multi-scale processing has emerged as a crucial component in brain tumor segmentation, addressing the inherent challenges of varying tumor sizes, shapes, and infiltration patterns. Pereira et al.\cite{pereira2016brain} demonstrated that effective multi-scale feature learning through a two-pathway CNN architecture could significantly improve the segmentation of both small and large tumor regions. Their approach systematically integrated different scales of information, showing particular effectiveness in handling heterogeneous tumor appearances.
Khened et al.\cite{khened2019fully} advanced multi-scale processing through a fully convolutional architecture with dense connectivity patterns, effectively combining features across multiple scales. Their method showed notable improvements in segmenting complex tumor boundaries by leveraging information from different resolutions, achieving competitive performance on the BraTS challenge dataset.
Islam et al.\cite{islam2020brain} introduced a multi-scale attention-guided network that effectively combined features at different resolutions through spatially-aware feature aggregation. Their approach demonstrated superior performance in capturing both fine-grained details and global context, particularly beneficial for accurate tumor boundary delineation in complex cases.

%% file: sections/emcad_methodology.tex
\section{EMCAD Methodology}

In this section, the EMCAD decoder is introuduced with the description of the two transformer-based architectures, PVT-EMCAD-B0 and PVT-EMCAD-B2. as seen in the work done by \cite{rahman2024emcad}.

\subsection{Efficient Multi-Scale Convolutional Attention Decoding (EMCAD)}
The EMCAD decoder processes multi-stage features extracted from pretrained hierarchical vision encoders for high-resolution semantic segmentation. EMCAD includes several components:

\begin{itemize}
    \item \textbf{Efficient Multi-Scale Convolutional Attention Modules (MSCAMs)} to enhance feature maps.
    \item \textbf{Large-Kernel Grouped Attention Gates (LGAGs)} that refine feature maps by merging them with skip connections via gated attention.
    \item \textbf{Efficient Up-Convolution Blocks (EUCBs)} for up-sampling, followed by feature enhancement.
    \item \textbf{Segmentation Heads (SHs)} at each stage to produce segmentation outputs.
\end{itemize}

The MSCAMs refine the pyramid features from the encoder's four stages, each followed by an SH that generates a segmentation map. The refined feature maps are then upscaled using EUCBs and combined with LGAG outputs. The model generates four segmentation maps, culminating in the final segmentation output.

\subsubsection{Multi-scale convolutional attention module (MSCAM)}

The Efficient Multi-Scale Convolutional Attention Module (MSCAM) refines feature maps using three components: a Channel Attention Block (CAB) to emphasize relevant channels, a Spatial Attention Block (SAB) to capture local context, and an Efficient Multi-Scale Convolution Block (MSCB) to enhance feature maps while preserving contextual relationships. By employing depth-wise convolution across multiple scales, MSCAM achieves high refinement efficiency at a considerably lower computational cost compared to traditional Convolutional Attention Modules (CAM).

The Multi-Scale Convolution Block (MSCB) enhances features generated by the cascaded expanding path, while the Channel Attention Block (CAB) assigns varying levels of importance to each channel, emphasizing relevant features and suppressing less relevant ones. Finally, the Spatial Attention Block (SAB) simulates the human brain's attentional processes, focusing on specific regions of the input image.

\subsubsection{ Large-kernel grouped attention gate (LGAG)}

A novel Large-Kernel Grouped Attention Gate (LGAG) is introduced to selectively enhance relevant feature maps by progressively combining them with attention coefficients. These coefficients, learned by the network, increase the activation of essential features while suppressing non-essential ones. A gating signal derived from higher-level features regulates information flow across network stages, thereby improving segmentation precision in medical imaging. 

\subsubsection{ Efficient up-convolution block (EUCB)}

The Efficient Up-Convolution Block (EUCB) is designed to upsample feature maps at each stage to align with the dimensions and resolution of the subsequent skip connection. EUCB first performs upsampling with a scale factor of 2, then enhances the upsampled features using a $3 \times 3$ depth-wise convolution followed by Batch Normalization (BN) and ReLU activation. A final $1 \times 1$ convolution reduces the number of channels to align with the next stage. The EUCB achieves high efficiency by employing depth-wise convolutions instead of traditional $3 \times 3$ convolutions.

\subsubsection{Segmentation head (SH)}
Segmentation Heads (SH) generate segmentation outputs from the refined feature maps across the four decoder stages. Each SH layer applies a $1 \times 1$ convolution to the refined feature maps, maintaining the same number of channels as in the given stage’s feature map. This operation produces an output with a number of channels matching the target dataset classes for multi-class segmentation, or a single channel for binary segmentation.

\subsection{EMCAD Architecture}

To demonstrate the generalization, effectiveness, and multi-scale feature processing capability of the EMCAD decoder in medical image segmentation, it is integrated with the PVTv2-B0 (Tiny) and PVTv2-B2 (Standard) networks \cite{wang2022pvt}. Unlike conventional transformer patch embeddings, PVTv2 enhances spatial information retention by incorporating convolutional operations. In the proposed PVT-EMCAD-B0 and PVT-EMCAD-B2 architectures, multi-scale features $(X_1, X_2, X_3, X_4)$ are extracted from four layers of the PVTv2 encoder. These features are fed into the EMCAD decoder, with $X_4$ entering the upsampling path and $X_3, X_2, X_1$ used in skip connections. EMCAD processes these inputs to produce segmentation maps corresponding to each encoder stage.

\begin{figure*}[!t] 
    \centering
    \includegraphics[width=\textwidth]{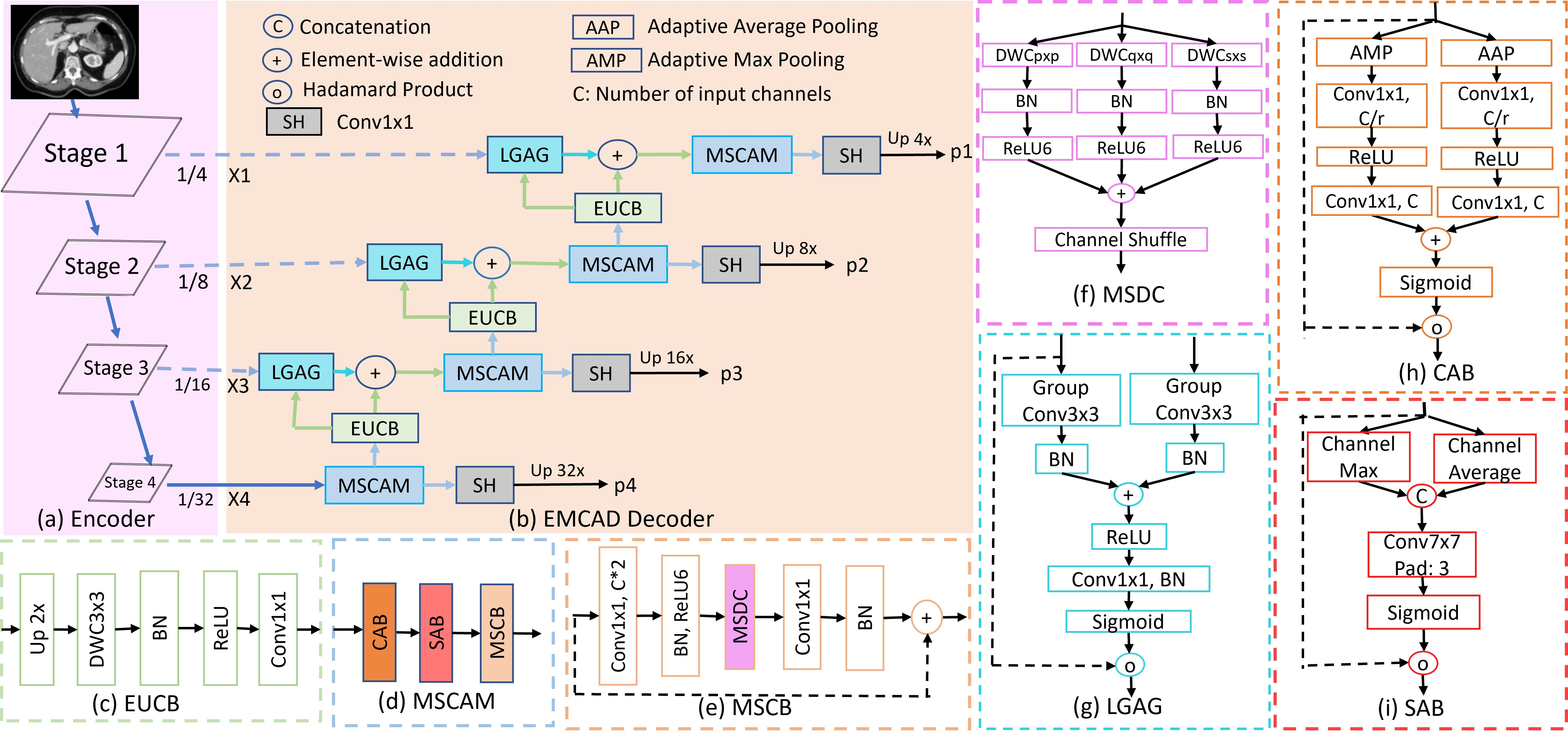}
    \caption{EMCAD Architecture \cite{rahman2024emcad}}
    \label{fig:emcad_architecture}
\end{figure*}

\subsection{Multi-stage loss and outputs aggregation}

The EMCAD decoder produces four prediction maps \( p_1, p_2, p_3, p_4 \) from its four segmentation heads. For loss aggregation, a combinatorial approach called MUTATION is adopted, inspired by \cite{rahman2024multi}, to combine the losses from all possible predictions derived from the four heads, yielding \( 2^4 - 1 = 15 \) unique predictions. The cumulative combinatorial loss is minimized during training. For binary segmentation, the additive loss is optimized with an additional term as in \cite{rahman2023medical}. For output aggregation, the final segmentation map is taken from the prediction map \( p_4 \) of the last stage, with a Sigmoid function used for binary segmentation and a Softmax function for multi-class segmentation.

%% file: sections/implementation.tex
\section{Implementation}

In this section, the details of the implementation of EMCAD on brain tumor segmentation is discussed. 

\subsection{Implementation Details}

The implementation was conducted on a Dell desktop with the following specifications: a 12th Gen Intel(R) Core(TM) i7-12700k 3.60GHz processor, 64GB RAM, and a 24.0 GB NVIDIA GeForce RTX 3090 GPU, running Windows 11 Professional 64-bit. The BRATS2020 \cite{menze2014multimodal} brain tumor segmentation dataset was used, containing scans from 369 patients with brain tumors. Each scan is represented as a 3D volume with dimensions of $240 \times 240 \times 155$, where there are 240 pixels in both the x and y directions and 155 slices along the z-axis. 

The dataset provides annotations for various tumor regions: 
\begin{itemize}
    \item GD-enhancing tumor (ET): Label 4
    \item Peritumoral edema (ED): Label 2
    \item Necrotic and non-enhancing tumor core (NCR/NET): Label 1
    \item Background: Label 0
\end{itemize}

To apply EMCAD for the BRATS2020 \cite{menze2014multimodal} dataset, several preprocessing steps were necessary. To begin, the dataset was split into 80\% for training and 20\% for testing. Data formats were adjusted to \texttt{.npz} for training and \texttt{.h5} for testing to accommodate model requirements. Next, images and their corresponding ground truths (GTs) were loaded. The normalization of the images was then performed by applying the ImageNet mean and standard deviation. This process adjusts the pixel values to have a mean of zero and a standard deviation of one, which helps in stabilizing the training process and improving the model's performance. 

It was also necessary to ensure that all images were in RGB format, a requirement for many deep-learning models. Finally, the ground truths were transformed to binary format (grayscale). This converts the GTs into a binary mask, where each pixel is either 0 or 1, indicating the absence or presence of a tumor. 

Pretrained ImageNet \cite{deng2009imagenet} PVTv2-b0 and PVT-b2 encoders \cite{wang2022pvt} were utilized. In the decoder’s Multi-Scale Convolutional Attention Module (MSCAM), kernel sizes were set to \{1, 3, 5\} as determined through an ablation study in \cite{rahman2024emcad}. Model training employed the AdamW optimizer \cite{loshchilov2017decoupled} with a learning rate and weight decay of $1e^{-4}$. Initial experiments involved training over 50 epochs with a batch size of 6 and a maximum of 50,000 iterations.

\subsection{Analysis of EMCAD Model Training and Evaluation Plots}

\begin{figure*}[!t]
    \centering
    \includegraphics[width=\textwidth]{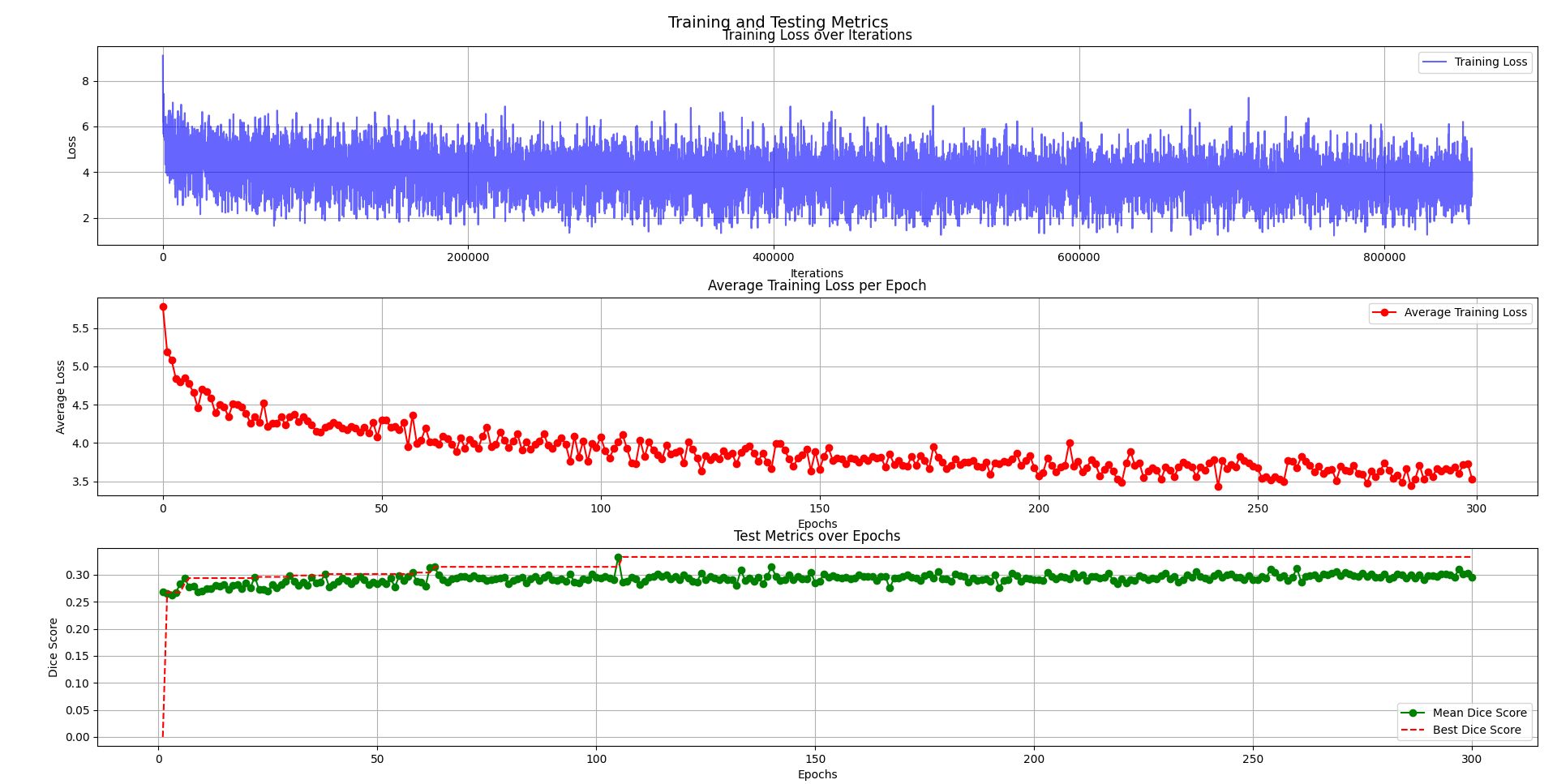}
    \caption{Training and validation metrics for batch size 6 and 300 epochs: (a) Per-iteration training loss, (b) Average training loss per epoch, (c) Test metrics including mean and best Dice scores.}
    \label{fig:prelim2}
\end{figure*}

\begin{figure*}[!t]
    \centering
    \includegraphics[width=\textwidth]{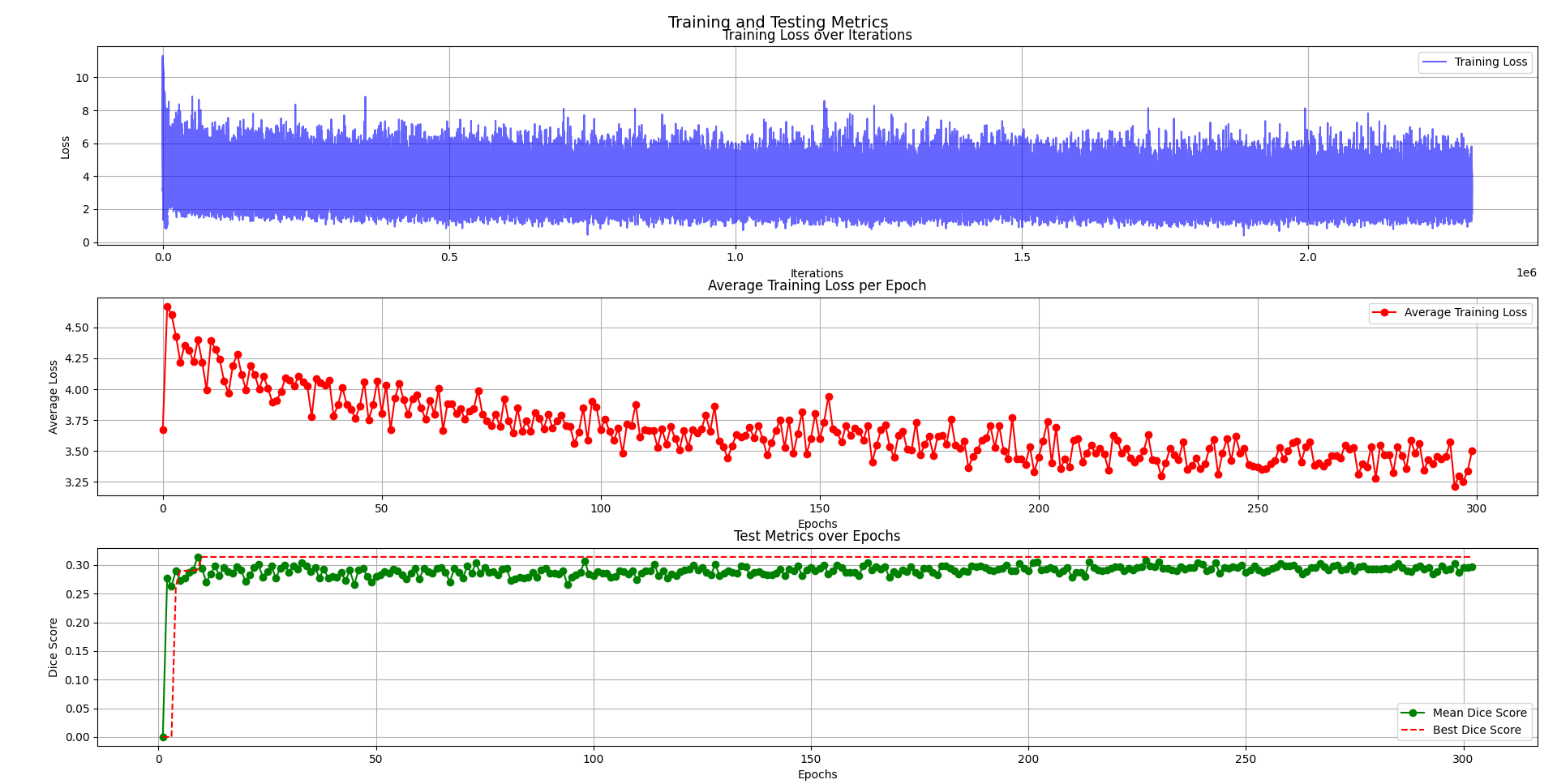}
    \caption{Training and validation metrics for batch size 16 and 300 epochs: (a) Per-iteration training loss, (b) Average training loss per epoch, (c) Test metrics including mean and best Dice scores.}
    \label{fig:prelim}
\end{figure*}

\begin{figure*}[!h]
    \centering
    \includegraphics[width=\textwidth]{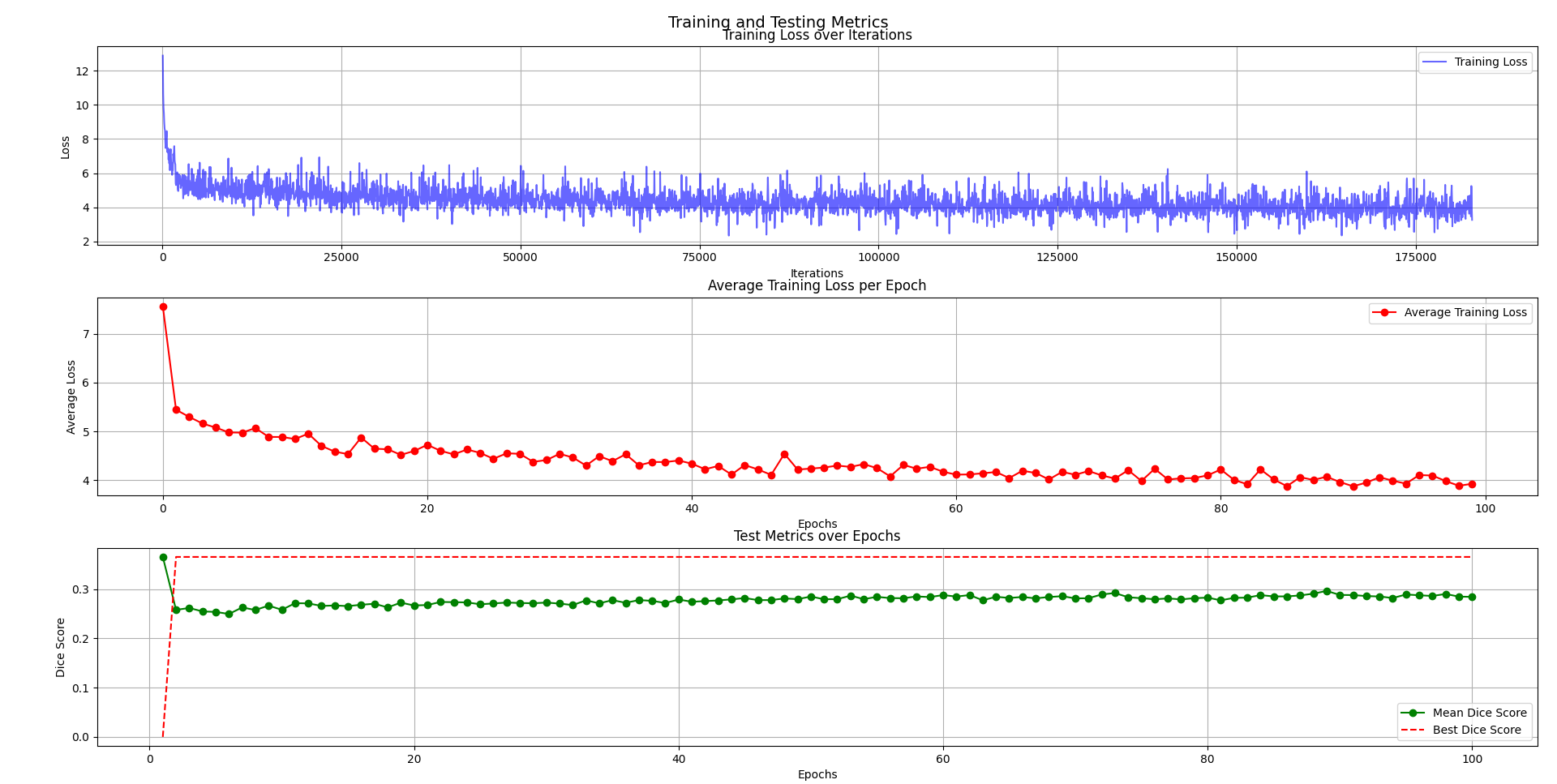}
    \caption{Training and validation metrics for batch size 25 and 100 epochs: (a) Per-iteration training loss, (b) Average training loss per epoch, (c) Test metrics including mean and best Dice scores.}
    \label{fig:prelim3}
\end{figure*}

Figures \ref{fig:prelim2}, \ref{fig:prelim} and \ref{fig:prelim3} show the training and testing metrics for our model. The top plot displays the training loss over iterations, the middle plot shows the average training loss per epoch, and the bottom plot illustrates the test metrics over epochs, specifically the mean and best Dice scores. We have extracted the data given by these figures into tables for better analysis.

\begin{table}[ht]
    \caption{Training Configuration Comparison}
    \centering
    \begin{tabular}{|l|c|c|c|}
        \hline
        \textbf{Parameter}          & \textbf{Batch Size 6} & \textbf{Batch Size 16} & \textbf{Batch Size 25} \\ \hline
        Epochs                      & 300                   & 300                    & 100                    \\ \hline
        Initial Loss                & $\sim$10              & $\sim$10               & $\sim$12               \\ \hline
        Final Avg Loss              & $\sim$3.5             & $\sim$3.5              & $\sim$4.0              \\ \hline
        Training Iterations         & $\sim$2.2M            & $\sim$800K             & $\sim$175K             \\ \hline
    \end{tabular}
    \label{table:training_config}
\end{table}

\begin{table}[ht]
    \caption{Performance Metrics}
    \centering
    \begin{tabular}{|l|c|c|c|}
        \hline
        \textbf{Metric}             & \textbf{Batch Size 6} & \textbf{Batch Size 16} & \textbf{Batch Size 25} \\ \hline
        Best Dice Score             & 0.310                 & 0.332                  & 0.365                  \\ \hline
        Mean Dice Score             & 0.285                 & 0.295                  & 0.284                  \\ \hline
        Score Stability             & Moderate              & High                   & High                   \\ \hline
        Convergence Speed           & Slow                  & Medium                 & Fast                   \\ \hline
    \end{tabular}
    \label{table:performance_metrics}
\end{table}

\begin{table}[ht]
    \caption{Training Loss Characteristics}
    \centering
    \begin{tabular}{|l|c|c|c|}
        \hline
        \textbf{Characteristic}     & \textbf{Batch Size 6} & \textbf{Batch Size 16} & \textbf{Batch Size 25} \\ \hline
        Loss Range                  & 2.0-8.0              & 2.0-6.0                & 2.0-6.0                \\ \hline
        Loss Volatility             & High                  & Medium                 & Low                    \\ \hline
        Convergence Pattern         & Gradual               & Smooth                 & Quick                  \\ \hline
        Loss Stability              & Unstable              & Stable                 & Very Stable            \\ \hline
    \end{tabular}
    \label{table:loss_characteristics}
\end{table}

\begin{table}[ht]
    \caption{Training Phases Analysis}
    \centering
    \begin{tabular}{|l|c|c|c|}
        \hline
        \textbf{Phase}              & \textbf{Batch Size 6} & \textbf{Batch Size 16} & \textbf{Batch Size 25} \\ \hline
        Initial Learning            & Fast                  & Fast                   & Very Fast              \\ \hline
        Mid-training Stability      & Poor                  & Good                   & Excellent              \\ \hline
        Final Convergence           & Incomplete            & Complete               & Complete               \\ \hline
        Loss Reduction Rate         & Slow                  & Medium                 & Fast                   \\ \hline
    \end{tabular}
    \label{table:phases_analysis}
\end{table}

\subsubsection{Training Configuration Comparison}
Table \ref{table:training_config} gives us the breakdown of the training efficiency of different batch sizes over the pre-determined number of epochs and iterations.

\textbf{Batch 6:} The model was trained for 300 epochs and required approximately 2.2M training iterations. The smaller batch size resulted in more frequent weight updates, leading to slower progress but a relatively decent comparable final average loss of around 3.5. This reflects the inefficiencies of smaller batch sizes for data processing and gradient stabilization.

\textbf{Batch 16:} For our batch size 16 experiment, the model trained for 300 epochs, but required only around 800K training iterations. Larger batches process more samples per step, providing more accurate gradient estimates, which explains the reduction in training iterations without sacrificing the final average loss.

\textbf{Batch 25:} Required the fewest iterations \( \sim 175K \) due to completing only 100 epochs. Despite this, the final average loss \( \sim 4.0 \) remained close to that of the other batch sizes, demonstrating efficient learning with fewer updates.

\subsubsection{Performance Metrics}
Table \ref{table:performance_metrics} measures the Dice scores and stability of the model at the end of training.

\textbf{Batch 6:} Achieved the lowest Best Dice Score (0.310) and Mean Dice Score (0.285). Furthermore, the moderate stability and slow progress it exhibited suggest that smaller batch sizes are less effective in optimizing segmentation performance.

\textbf{Batch 16:} Showed improved Best Dice Score (0.332) and Mean Dice Score (0.295), with high stability as well as medium convergence speed. This indicates better gradient estimation and optimization due to the larger batch size.

\textbf{Batch 25:} Achieved the highest Best Dice Score (0.365) but slightly lower Mean Dice Score (0.284). The higher peak score suggests that the model benefited from stable training dynamics, even with a shorter training schedule.

\subsubsection{Training Loss Characteristics}
Table \ref{table:loss_characteristics} analyzes loss behavior across the training period.

\textbf{Batch 6:} Exhibited high volatility and a broad loss range (2.0-8.0), indicating unstable training dynamics. The gradual progress suggests frequent weight updates introduce noise, therefore delaying consistent improvement.

\textbf{Batch 16:} Reduced volatility and loss range (2.0-6.0), while keeping a more consistent progress pattern. This reflects improved training stability because of the greater sample size every iteration.

\textbf{Batch 25:} Demonstrated the most consistent loss behavior, with the quickest improvement and the lowest volatility. The highly stable loss demonstrates the effectiveness of greater batch sizes in stabilizing gradient updates.

\subsubsection{Training Phases Analysis}
Table \ref{table:phases_analysis} illustrates learning dynamics across different training phases.

\textbf{Batch 6:} Demonstrated fast initial learning but poor mid-training stability and incomplete reduction in loss. The slower progress reflects difficulties in minimizing loss due to noisy gradients.

\textbf{Batch 16:} Achieved a balance across phases, with good mid-training stability and steady progress toward reducing loss.

\textbf{Batch 25:} Excelled across all phases, showing speedy initial learning, excellent mid-training stability, and the fastest reduction in loss.

\textbf{Key Observations:}
\begin{itemize}
    \item \textbf{Progress Acceleration with Larger Batch Sizes:} Larger batch sizes (16 and 25) demonstrated faster and more stable progress within the fixed number of iterations and epochs. Batch Size 25 achieved the highest Dice score (0.365) within the shortest schedule, highlighting its efficiency in leveraging larger updates per iteration.
    \item \textbf{Stability Advantages:} Larger batch sizes reduced gradient noise, leading to smoother and more stable training loss. This stability enabled better performance metrics, especially for Batch Sizes 16 and 25, compared to Batch Size 6.
    \item \textbf{Efficiency Gains:} Batch Size 25 processed more data per iteration, achieving comparable or superior performance within significantly fewer epochs and iterations. However, the memory overhead associated with larger batch sizes must be considered.
    \item \textbf{Trade-offs:} While smaller batch sizes require more updates to progress, they are computationally lighter. Larger batch sizes excel in stability and efficiency but demand more computational resources, making Batch Size 16 a good compromise for systems with limited memory.
    \item \textbf{Learning Dynamics:} Across all training phases, Batch Size 25 made the fastest progress, with excellent stability and loss reduction, emphasizing its capability to reach higher performance levels in shorter training schedules.
\end{itemize}

%% file: sections/future-work.tex
\section{Future Work}


To improve the performance and stability of the current model, the following strategies are proposed:

\begin{itemize}
    \item \textbf{Adaptive Learning Rate Scheduling:} Implement learning rate scheduling techniques to improve stability and training progress, such as cyclical learning rates or warm restarts.
    \item \textbf{Further Exploration of Batch Sizes:} Test larger batch sizes, such as 18, 20, or 30, and explore very large batch sizes combined with gradient accumulation to improve stability and efficiency.
    \item \textbf{Mixed Precision Training:} Utilize mixed precision training to reduce memory usage and allow for larger batch sizes or deeper architectures while maintaining model accuracy.
    \item \textbf{Enhanced Regularization Techniques:} Apply spatial dropout and path dropout to improve generalization performance and reduce overfitting.
    \item \textbf{Model Architecture Enhancements:} Investigate deeper architectures and memory-efficient layers to enhance model capacity and efficiency.
    \item \textbf{Data Augmentation and Normalization:} Incorporate advanced data augmentation methods and improved normalization to enhance input data consistency and robustness.
    \item \textbf{Gradient Clipping:} Implement gradient clipping to stabilize training and prevent extreme gradient updates, especially for smaller batch sizes.
    \item \textbf{Optimizer Fine-Tuning:} Fine-tune the optimizer and experiment with alternatives like AdamW or RAdam for better stability and convergence.
\end{itemize}

\section{Expected Outcomes}

The implementation of these optimizations is expected to:
\begin{itemize}
    \item \textbf{Improved Stability:} Reduce loss volatility and produce smoother training curves.
    \item \textbf{Better Performance:} Achieve higher Dice scores and improved segmentation quality.
    \item \textbf{Training Efficiency:} Reduce the number of iterations required, optimizing memory and computational efficiency.
    \item \textbf{Enhanced Generalization:} Improve robustness, narrowing the gap between training and validation performance.
    \item \textbf{Scalability:} Enable better adaptation to different computational setups and support experimentation with larger datasets or architectures.
\end{itemize}

%% file: sections/conclusion.tex
\section{Conclusion}

In this study, the EMCAD model was applied to the BraTS 2020 brain tumor dataset, comprising MRI scans from 369 patients, to evaluate its effectiveness in medical image segmentation. Experiments showed us that while EMCAD demonstrates reliable generalization across cases, further optimizations could enhance training stability and convergence, potentially improving segmentation accuracy. Adjustments such as adaptive learning rate scheduling, further exploration of larger batch sizes, and enhanced regularization techniques may reduce loss volatility and support model performance in capturing complex tumor structures. These refinements could strengthen EMCAD's applicability for robust, high-accuracy segmentation of brain tumors and other clinical settings.

However, it is also worth noting that exploring other models for brain tumor segmentation may yield better results, as EMCAD may not be the ideal model for this task. Further exploration of alternative architectures could lead to more effective solutions for accurate segmentation of brain tumors. Also, incorporating advanced techniques such as data augmentation, transfer learning, and ensemble methods could further enhance the model's performance. Collaborating with domain experts to fine-tune the model parameters and training with larger, more diverse datasets could also contribute to achieving more accurate and reliable segmentation results. Ultimately, a comprehensive evaluation of different models is needed to identify which is the most suitable solution for brain tumor segmentation, thereby ensuring the best possible outcomes for real-world applications.